\documentclass{nature}

\usepackage{amssymb}
\usepackage{graphicx}
\usepackage{times}
\usepackage{epsfig}
\usepackage{amsmath}
\usepackage{subfigure}
\usepackage{bm}

\usepackage{mathtools}
\usepackage{epstopdf}
\usepackage[font=scriptsize]{caption}
\usepackage{float}

\usepackage{algorithm}
\usepackage[noend]{algorithmic}

\newcommand{\argmin}{\arg\!\min}
\title{A Survey of Efficient Regression of General-Activity Human Poses from Depth Images}

\author{Wenye He}

\begin{document}

\maketitle

\begin{abstract}
This paper presents a comprehensive review on regression-based method for human pose estimation. The problem of human pose estimation has been intensively studied and enabled many application from entertainment to training.  Traditional methods often rely on color image only which cannot completely ambiguity of joint's 3D position, especially in the complex context.  With the popularity of depth sensors, the precision of 3D estimation has significant improvement. In this paper, we give a detailed analysis of state-of-the-art on  human pose estimation, including depth image based and RGB-D based approaches. The experimental results demonstrate their advantages and limitation for different scenarios. 
\end{abstract}

\section*{Introduction}
Human pose estimation from images has been studied for decades in computer vision. As recent development in cameras and sensors, depth images receive a wide spread of notice from researchers from body pose estimation \cite{iref1} to 3D reconstruction \cite{ierf1}. Girshick et al.\cite{iref1} present an approach to find the joints position in human body from depth images. They address the problem of general-activity pose estimation. Their regression-based approach sucessfully computes the joint positions even with occlusion. Their method can be view as a new combination of two existing works, implicit shape models\cite{iref2} and Hough forest\cite{iref3}. The following sections cover related works, explanation on the method from testing to training, and result and comparison.

\section*{Related Works}
In previous works, one common idea in human pose estimation is to focus on finding different body parts. Bourdev et al.\cite{iref4} put foward a two-layer regression model that trains segments classifers to local patterns detection and combines the output of the classifiers. Plagemann et al.\cite{iref5} create a novel interest point detector for catching body components from depth images. Shotton et al.\cite{iref6} design a system that change an single input depth image to an inferred per-pixel body part distribution and local the 3D joint position. Pictorial structures based methods have also been used for human pose estimation to enhance the body shape \cite{iref7}. This method can optimally remove the ambiguity of 3D inference from a single view point. Pictorial structures can also be used together with segmentation to efficiently localize body part and predict the joint position \cite{iref8}. However, this idea has some problems with the consideration of the required definition of body alignment, joints inside the body, and body occlusion. The implicit shape model\cite{iref9} can solve these problems. Random forest\cite{iref10} based method also have advantages over the body-part-based methods.

\section*{Method}
The researchers call their approach Joint Position Regression. Their algorithm find joint points of 3D human body from aggregated votes from a regression forest. The testing and training procedures are introduced as follow.\\
A regression forest is made up from a group of decision trees which give the predicted outputs. Every tree is binary and contains split nodes and leaf nodes. The split nodes have tests. In the test, the researchers compute the feature value by comparing the depth at nearby pixels to a threshold. The result of each test determines whether to branch to the left or right child. The leaf nodes are the end of a tree. Given different inputs into the root, the inputs go through a sequence of corresponding test in split nodes of each depth and finally return the corresponding output in the leaf node. The researchers store a few relative votes at each leaf node. They define the set of relative votes for joint $j$ at node $l$ as $V_{lj}=\{(\delta_{ljk},w_{ljk})\}^K_{k=1}$, where $\delta_{ljk}\in\mathbb{R}^3$ is a 3D relative vote vectors, $w_{ljk}$ is a confidence weight to each vote, and $K$ is the number of votes in each leaf node. The vectors are obtained by taking the centers of the $K$ largest modes, the most frequently appeared value, found by mean shift. The weights are given by the sizes of their cluster. $K$ is kept to be small for efficiency, e.g.$K=1$ or $2$.
The testing algorithm is showed in Algorithm 1 and Aggregation of pixel votes at test time is showed in Figure 1. \begin{algorithm}
\begin{spacing}{1.0}
\caption{Inferring joint position hypotheses}
\begin{algorithmic}
\STATE // Collect absolute votes
\STATE initialize $Z_j=\emptyset$ for all joints $j$
\FORALL{pixels $q$ in the test image}
\STATE lookup 3D pixel position $\mathbf{x}_q=(x_q,y_q,z_q)^{\top}$
\FORALL{trees in forest}
\STATE descend tree to reach leaf node $l$
\FORALL{joints $j$}
\STATE lookup weighted relative vote set $V_{lj}$
\FORALL{$(\Delta_{ljk},w_{ljk})\in V_{lj}$}
\IF{$||\Delta_{ljk}||_2\leq distance threshold \lambda_j$}
\STATE compute absolute vote $\mathbf{z}=\Delta_{ljk}+\mathbf{x}_q$
\STATE adapt confidence weight $w=w_{ljk}\cdot z^2_q$
\STATE $Z_j:=Z_j\cup\{(\mathbf{z},w)\}$
\ENDIF
\ENDFOR
\ENDFOR
\ENDFOR
\ENDFOR
\STATE //Aggregate weighted votes
\STATE sub-sample $Z_j$ to contain $N$ votes
\STATE aggregate $Z_j$ using mean shift on Eq.1
\RETURN weighted nodes as final hypotheses
\end{algorithmic}
\end{spacing}
\end{algorithm}
\begin{figure*}[h]
\centerline{\epsfig{figure=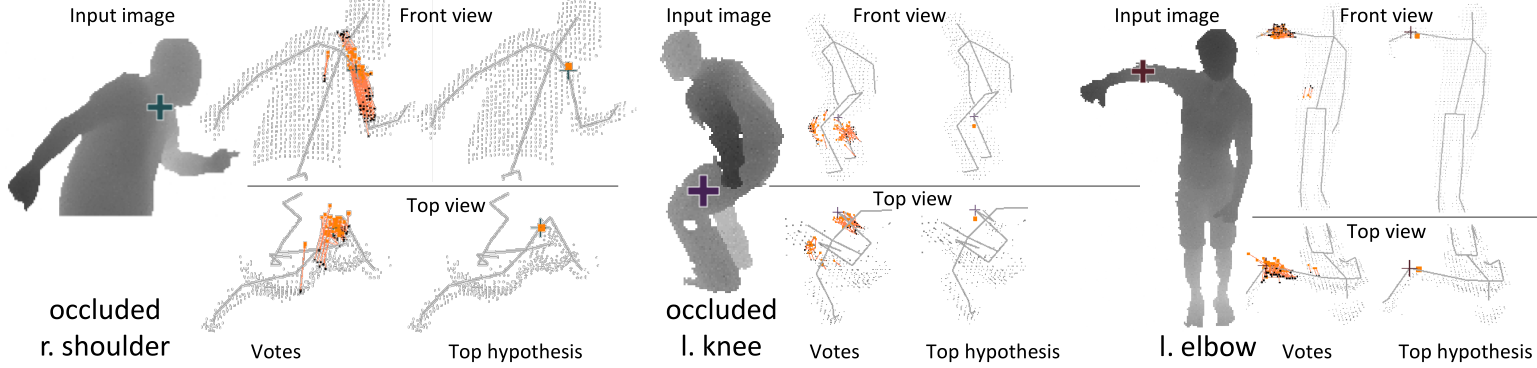,width=18cm}}
\caption{Each pixel (black square) casts a 3 D vote (orange line) for each joint. Mean shift is used to aggregate these votes and produce a final set of hypotheses for each joint. Note accurate predictions of internal body joints even when occluded. The highest confidence hypothesis for each joint is shown.}
\end{figure*}
The set $Z_j$ of abosolute votes cast by all pixels for each body joint $j$.\\
The algorithm include 3 steps, collecting absolute votes, aggregate weighted votes and computing final hypotheses. The set of absolute votes is updated by adding the 3D pixel position into the reliable weighted relative vote with the adapt confidence weight. The threshold $\lambda_j$ in the algorithm are used for elimation of unreliable predictions. Finally, the algorithm returns aggregated using mean shift using Eq. 1.\begin{equation}
p_j(\mathbf{z}')\propto\sum_{(\mathbf{z},w)\in Z_j}w\cdot exp(-||\frac{\mathbf{z}'-\mathbf{z}}{b_j}||^2_2),
\end{equation}
where $b_j$ is a learned per-joint bandwidth and $\mathbf{z}'$ is world space. Figure show aggregation of pixel votes in testing.\\
After the explanation of the testing procedure, the training is demonstrated here. Training is composed of three learning, the leaf node regression models, the hyper-parameters and the tree structure.\\
In the first learning, the major objective is to learn the set of relative votes. Algorithm 2 below show how to achieve the goal. \begin{algorithm}
\begin{spacing}{1.0}
\caption{Learning relative votes}
\begin{algorithmic}
\STATE // Collect relative offsets
\STATE initialize $R_{lj}=\emptyset$ for all leaf nodes $l$ and joints $j$
\FORALL{pixels $q$ in all training images $i$}
\STATE lookup ground truth joint positions $\mathbf{z}_{ij}$
\STATE lookup 3D pixel position $\mathbf{x}_{iq}$
\STATE compute relative offset $\Delta_{iq\to j}\mathbf{z}_{ij}-\mathbf{x}_{iq}$
\STATE descend tree to reach leaf node $l$
\STATE store $\Delta_{iq\to j}$ in $R_{lj}$ with reservoir sampling
\ENDFOR
\STATE // Cluster
\FORALL{leaf nodes $l$ and joints $j$}
\STATE cluster offsets $R_{lj}$ using mean shift
\STATE take top $K$ weighted modes as $V_{lj}$
\ENDFOR
\RETURN relative votes $V_{lj}$ for all nodes and joints
\end{algorithmic}
\end{spacing}
\end{algorithm}The simple process of computing relative votes $V_{lj}$ is that find the differences between the ground truth joint position and the 3D pixel position, throw the differences into different group using mean shift and pick the best $K$ relative votes.
The goal of the second learning is to find the optimal bandwidth and thresholds for this method. The researchers find the bandwidth $b^* = 0.005$m and the threshold $\lambda_j$ fall between $0.1$m and $0.55$m.
In the third learning, the researchers use the standard greedy decision tree. They use Eq.2 to repeat splitting the set of all training pixels $Q=\{(i,q)\}$ into left $Q_l(\phi)$ and right $Q_r(\phi)$ subsets.\begin{equation}
\phi^*=\argmin_\phi\sum_{s\in\{l,r\}}\frac{|Q_s(\phi)|}{|Q|}E(Q_s(\phi))
\end{equation}
$E(Q)$ is an error function to reducing the error in the partitions. The researchers use both regression error function, $E^{reg}(Q)$, and classification error one, $E^{cls}(Q)$, and observe the different result. For regression, they apply the method purposed by, while, for classification, they employ the method presented by.

\section*{Experiments and results}
In this section, the researchers evaluated their tree structures and compare their work with existing work. They evaluate their method on the MSRC\cite{iref6} dataset. To measure accuracy, they compare average precision and mean across joints (mAP) in experiments. They use forests of 3 trees each of which are trained to depth 20 with 5000 images. Figure 2 show some example of joints inferences in the researchers' method.\\
\begin{figure*}[h]
\centerline{\epsfig{figure=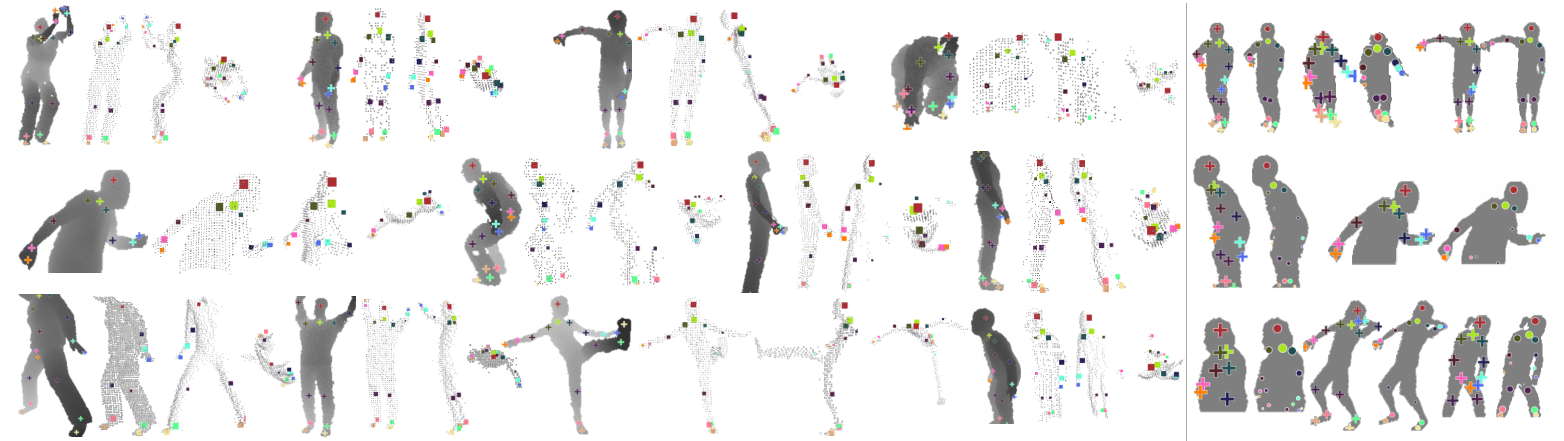,width=18cm}}
\caption{In the left group, each example shows an input depth image with colored ground truth joint positions, and inferred joint positions from front, right, and top views. The size of the boxes implies the inferred confidence. In the right group, example inference results on flattened 2D silhouettes. The crosses are the ground truth joint positions and the circles with size indicating confidence are the highest scoring hypothesis.}
\end{figure*}In the end of the previous section, regression and classification objective functions are mentioned. Figure 3 show average precision of them on all joints. Classification objective function gives the highest accuracy, so it is used in later experiments for comparisons.\begin{figure*}[h]
\centerline{\epsfig{figure=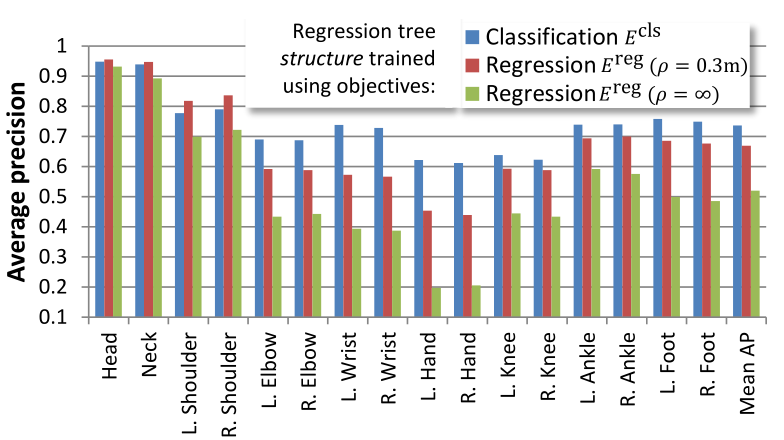,width=12cm}}
\caption{Comparison of tree structure training objectives. $\rho$ is the thresholds used in regression objective functions.}
\end{figure*}
\subsection{Hough forests}
Figure 4 show the result of comparisons. The researchers use Hough forests on MSRC-5000 test data with different votes and tree structures.\begin{figure*}[h]
\centerline{\epsfig{figure=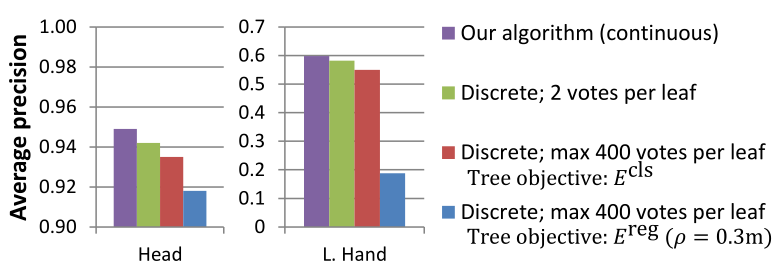,width=12cm}}
\caption{Comparison with Hough forest. $\rho$ is the thresholds used in regression objective functions. Different votes and objective functions are used in Hough forest.}
\end{figure*}
\subsection{Shotton et al.\cite{iref6}}
Figure 5 show the result of comparisons. The researchers’algorithm get higher mAP than Shotton et al.’s one with different sizes of training set.\begin{figure*}[h]
\centerline{\epsfig{figure=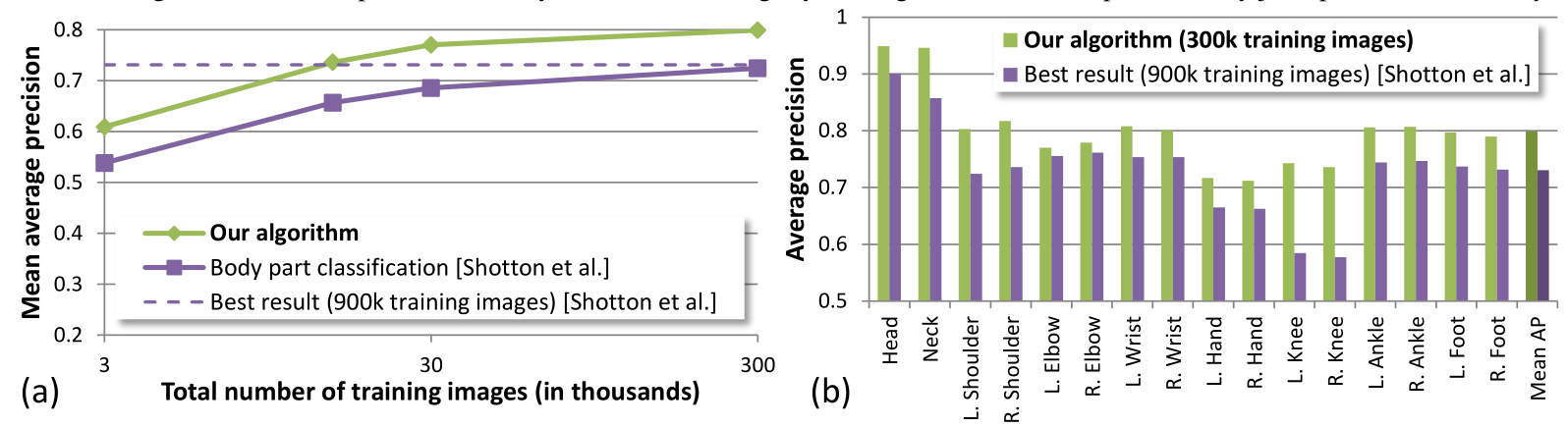,width=18cm}}
\caption{Comparison with Shotton et al. (a) Mean average precision versus total number of training images. (b) Average precision on each of the 16 test body joints.}
\end{figure*}


\section*{Reference}

\begin{thebibliography}{1}

\bibitem{iref1} Girshick, Ross, et al. "Efficient regression of general-activity human poses from depth images." Computer Vision (ICCV), 2011 IEEE International Conference on. IEEE, 2011.

\bibitem{ierf1} J. Shen and S. C. S. Cheung, ``Layer Depth Denoising and Completion for Structured-Light RGB-D Cameras," IEEE Conference on Computer Vision and Pattern Recognition,  pp. 1187-1194, 2013.

\bibitem{iref2} Müller, Jürgen, and Michael Arens. "Human pose estimation with implicit shape models." Proceedings of the first ACM international workshop on Analysis and retrieval of tracked events and motion in imagery streams. ACM, 2010.

\bibitem{iref3} Gall, Juergen, and Victor Lempitsky. "Class-specific hough forests for object detection." Decision forests for computer vision and medical image analysis. Springer London, 2013. 143-157.


\bibitem{iref4} Bourdev, Lubomir, and Jitendra Malik. "Poselets: Body part detectors trained using 3d human pose annotations." Computer Vision, 2009 IEEE 12th International Conference on. IEEE, 2009.



\bibitem{iref5} Plagemann, Christian, et al. "Real-time identification and localization of body parts from depth images." Robotics and Automation (ICRA), 2010 IEEE International Conference on. IEEE, 2010.

\bibitem{iref6} Shotton, Jamie, et al. "Real-time human pose recognition in parts from single depth images." Communications of the ACM 56.1 (2013): 116-124.


\bibitem{iref7} J. Shen and J. Yang, ``Automatic human animation for non-humanoid 3d characters," International Conference on Computer-Aided Design and Computer Graphics (CAD/Graphics),  pp. 220-221, 2015.

\bibitem{iref8} J. Shen and Y. Yang ``Automatic pose tracking and motion transfer to arbitrary 3d characters," International Conference on Image and Graphics,  pp. 640-653, 2015.

\bibitem{iref9} Leibe, Bastian, Aleš Leonardis, and Bernt Schiele. "Robust object detection with interleaved categorization and segmentation." International journal of computer vision 77.1-3 (2008): 259-289.

\bibitem{iref10} Rogez, Grégory, et al. "Randomized trees for human pose detection." Computer Vision and Pattern Recognition, 2008. CVPR 2008. IEEE Conference on. IEEE, 2008.




\end{thebibliography}
\end{document}